\documentclass[12pt]{article}
\usepackage[utf8]{inputenc}
\usepackage[T1]{fontenc}
\usepackage{amsmath,amsfonts,amssymb}
\usepackage{graphicx}
\usepackage{a4wide}

\begin{document}
\title{Additive regularization schedule for neural architecture search}
\author{Mark Potanin,  Kirill Vayser, and Vadim Strijov\footnote{E-mail: vadim@m1p.org}}
\maketitle

\begin{abstract}
Neural network structures have a critical impact on the accuracy and stability of forecasting. Neural architecture search procedures help design an optimal neural network according to some loss function, which represents a set of quality criteria. This paper investigates the problem of neural network structure optimization. It proposes a way to construct a loss function, which contains a set of additive elements. Each element is called the regularizer. It corresponds to some part of the neural network structure and represents a criterion to optimize. The optimization procedure changes the structure in iterations. To optimize various parts of the structure, the procedure changes the set of regularizers according to some schedule. The authors propose a way to construct the additive regularization schedule. By comparing regularized models with non-regularized ones for a collection of datasets the computational experiments show that the proposed method finds efficient neural network structure and delivers accurate networks of low complexity.
\end{abstract}
\noindent \textbf{Keywords:} machine learning; data science; stochastic processes; genetic algorithms; neural architecture search; additive regularization
 
\section{Introduction}
A deep learning neural network of optimal structure approximates data as accurately as possible. It  models the initial unknown dependence of the target variable~$y$ given vectors~$\mathbf{x}$ of the feature space. The model maps the input~$\mathbf{x}$ to the target~$y$ as \mbox{$f:(\mathbf{x},\mathbf{W})\mapsto y$.} In general, the deep learning model is a superposition of neurons: generalized linear models and activation functions. It can contain a large number of layers and neurons. It is assumed that the more complex the model, the higher its approximation accuracy, that is, the lower the value of the error function. The paper~\cite{KRAUS2020628} enlightens the influence of the method of constructing the error function on the choice of the structure of the deep learning model. 

We have to develop a method to reduce the complexity of the model while maintaining its accuracy. The quality function consists of an approximation loss function and a set of additive regularization elements to optimize the model structure and parameters. These additive elements are weighted, so that the quality function is treated as some linear combination. The authors proposed a method, which set a schedule of the model structure optimization. During the optimization procedure, the error function includes and excludes certain regularization elements to choose the optimal structure.

\emph{Regularization} is a method to optimize the model parameters. It includes to the loss function additional elements called regularizers~$\mathcal{R}_i$. Each element takes into account the specific requirements of the problem being solved to modify the structure and the parameters of the model during the optimization procedure. Regularization increases the stability of the forecast in the case of multiple correlation of model parameters or variables of the feature space. It promotes increasing the generalizing ability of the model and reducing the risk of overfitting ~\cite{svensen2007pattern}. \emph{Additive regularization} is a kind of regularization based on optimization of the weighted sum of regularizers.

The object of our research is a method of the error function constructing. We study the influence of the weights of regularizes~(1) on the complexity and accuracy of the model. The additive regularization function~\cite{vorontsov2015additive} is
\[
L= \sum_{i=1}^r \lambda_i\mathcal{R}_i,
\]
where~$\mathcal{R}_i$ is a regularizer. It modifies parameters of the model during optimization. Call the \emph{metaparameters} the weights of the regularizers~$\lambda_i$.
\paragraph{Method.} Values of the metaparameters~$\lambda_i$ are changed in the run of the structure optimization procedure. This change is called the~\emph{optimization schedule}. Unlike the regularization for linear models, metaparameters of additive regularization are assigned to each of the layers of the neural network. For example, an element of an error function with additive regularization~$\mathcal{R}$ is
\[
\lambda\|\mathbf{w}\|_2^2,
\]
where~$\lambda$  is the metaparameter of the error function, and~$\mathbf{w}$ is the model parameters.

Varying the values of metaparameters adds or removes from consideration individual terms of additive regularization, and changes their impact on the error function. Individual regularizers for each layer of the neural network are added to the error function, which is added sequentially. Our search for a model structure works through the addition or removal of elements from the structure of a neural network using a genetic algorithm. We call it GA-NAS (Genetic Algorithm for Neural Architecture Search). The \emph{structural complexity} of a neural network is the number of model parameters.

\begin{figure}[!htbp] 
\centering\includegraphics[width=\textwidth]{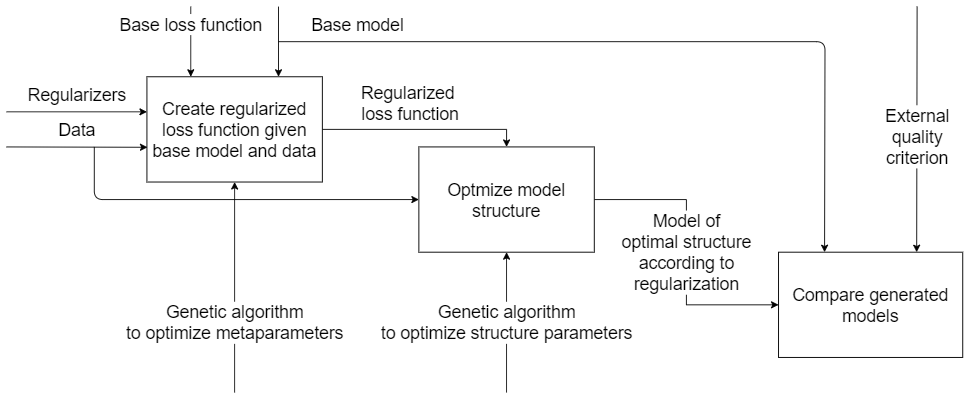}
\caption{Functional diagram of the proposed method}\label{fig:diag}
\end{figure}

The model optimization algorithm consists of the following steps: 1) setting an initial structure of the deep learning model, 2) optimizing the model parameters, 3) optimizing the metaparameters, 4) pruning the structure with GA-NAS. Figure~\ref{fig:diag} shows the scheme of the proposed algorithm. The model parameters are optimized by a backpropagation algorithm. The metaparameters are optimized using a genetic algorithm. We call this method GA-REG (Genetic Additive REGularization). The research paper has proposed a method for the simultaneous optimization of the model parameters and metaparameters of additive regularization.

\paragraph{Related work.} Tikhonov regularization solves the problem of multicollinearity in linear regression, which commonly occurs in models with large numbers of parameters. The~$L_2$-regularization is described in~\cite{tikhonov1965}. It works by balancing bias and variance. Its disadvantage is that it cannot create a sparse model, since it retains the entire set of parameters. Another type~$L_1$ of the regularization was proposed in~\cite{tibshirani1996regression}, and it leads to the automatic selection of parameters. It also has several disadvantages, including:
\begin{enumerate}
\item If we denote by~$p$ the number of independent variables and by~$n$ the number of samples in the dataset, then in the case~$p > n$ the Lasso regularization chooses maximum~$n$ of independent variables from the set.
\item In the presence of groups of strongly correlated variables, the Lasso regularization chooses only one variable from the group, not paying attention to which one.
\item In the case of~$n > p$ and a high correlation between the variables, it has been empirically shown that ridge regression performs much better than Lasso.
\end{enumerate}
Thus, items 1) and 2) make Lasso an inapplicable technique in some problems where feature selection is required. The listed problems are solved using another technique of elastic net regularization~\cite{zou2005regularization}. It allows making automatic selection of variables, 
adjusting their weights, as well as choosing groups of correlating features. The elastic net regularization method is the addition of two additional terms to the error function
\[
L\left(\lambda_{1}, \lambda_{2}, \lambda_{3}, \mathbf{w}\right)=\lambda_{1}\|\mathbf{y}-\mathbf{X} \mathbf{w}\|^{2}+\lambda_{2}\|\mathbf{w}\|^{2}+\lambda_{3}\|\mathbf{w}\|_{1}.
\]

A generalization of Elastic Net is the Support Features Machine~(SFM) regularization~\cite{tatarchuk2010support}. The error function is
\begin{gather*} 
\min _{\mathbf{w}} S(\mathbf{w})=C \sum_{i=1}^{l}\bigl(1-M_{i}(\mathbf{W})\bigr)_{+}+\sum_{j=1}^{n} \mathcal{R}_{\mu}\left(w_{j}\right),  \tag{2}\\
\mathcal{R}_{\mu}\left(w_{j}\right)= \begin{cases}2 \mu\left|w_{j}\right|, & \left|w_{j}\right| \leqslant \mu, \\
\mu^{2}+w_{j}^{2}, & \left|w_{j}\right| \geqslant \mu.\end{cases}
\end{gather*}
The features are chosen using the selectivity parameter~$\mu$. Noise features~$\left(\left|w_{j}\right|<\mu\right)$ are suppressed as in Lasso, significant dependent features are grouped as in the elastic net. A method like Relevance Features Machine~(RFM) is similar to it. It is given as follows:
\begin{equation*}
\min _{\mathbf{w}} C \sum_{i=1}^{l}\bigl(1-M_{i}(\mathbf{w})\bigr)_{+}+\sum_{j=1}^{n} \ln \left(w_{j}^{2}+\frac{1}{\mu}\right). \tag{3}
\end{equation*}
RFM~(3) shows more effective feature selection than SFM~(2).

In general, if the function~$L$ is convex, then the Moreau-Yosida regularization can be used. We write~$\lambda>0$
\[
M_{\lambda L}(\mathbf{x})=\inf _{\mathbf{u}}\left(L(\mathbf{u})+\frac{1}{2 \lambda}\|\mathbf{x}-\mathbf{u}\|_{2}^{2}\right).
\]
This function has remarkable properties: 1)~$M_{\lambda f}(\mathbf{x})$ is a convex function by virtue of infimal convolution; 2) the set of minimum points for~$L$ and~$M_{\lambda L}(\mathbf{x})$ coincide; 3)~$M_{\lambda L}(\mathbf{x})$ is a smooth function due to the strong convexity of its first conjugate function and coincidence with  second conjugate function.%~$M_{\lambda L}(\mathbf{x})=M_{\lambda L}^{* *}(\mathbf{x})$.

The paper~\cite{Haber_2017}  presents several strategies for stabilizing the optimization of the structure of deep networks. This article also provides a description of the regularization that is sensitive to a smooth change in model parameters between adjacent layers. In addition, as shown in numerical experiments, the regularization increases the stability of the model.

For the practical use of additive regularization, it is important that the model is interpretable in addition to the high accuracy of approximation. In~\cite{wang2018learning}, the EYE regularization~(expert yielded estimates) is considered. It includes expert knowledge of the relationship between traits and the dependent variable. The problem of minimizing empirical risk is 
\[
\hat{\mathbf{W}}=\min S(\mathbf{W}, \mathbf{X}, \mathbf{y})+n \lambda L(\mathbf{W}, \boldsymbol{\Gamma}).
\]
It minimizes the sum of the error function and the regularizer~$L$. There are many~$\mathcal{D}$ features, of which for the set where~$\mathcal{K} \subseteq \mathcal{D}$, there is additional information that these features are important in the considered expert area. Therefore, sparseness is required for~$\hat{\mathbf{W}}_{\mathcal{D} \backslash \mathcal{K}}$, but not for~$\hat{\mathbf{W}}_{\mathcal{K}}$. Our basic approach is to use the~$L_{1}$ and~$L_{2}$ regularization. Their regularization term is
\[
L=(1-\beta)\|\boldsymbol{\Gamma} \odot \mathbf{W}\|_{2}^{2}+\beta\|(1-\boldsymbol{\Gamma}) \odot \mathbf{W}\|_{1}, 
\]
where the parameter~$\beta$ controls the balance between the features from~$\mathcal{K}$ and~$\mathcal{D} \backslash \mathcal{K}$. The solution proposed in~\cite{Haber_2017} is
\begin{equation*}
L=\|(1-\boldsymbol{\Gamma}) \odot \mathbf{W}\|_{1}+\sqrt{\|(1-\boldsymbol{\Gamma}) \odot \mathbf{W}\|_{1}^{2}+\|\boldsymbol{\Gamma} \odot \mathbf{W}\|_{2}^{2}}. \tag{4}
\end{equation*}
It made possible to obtain models in which the main features used strongly overlap with factors that were considered significant in the medical environment in problems of patient risk stratification. Also, it maintains high prediction quality.

Some other methods of neural network regularization are: data augmentation~\cite{krizhevsky2012imagenet} reduces the approximation error, batch normalization~\cite{KRAUS2020628}
stabilizes the model, and dropout~\cite{srivastava2014dropout} prevents overfitting. 

Many regularization methods possess a significant disadvantage: metaparameters are constant throughout the entire learning process. This research paper proposes an approach that makes it possible to change the influence of additive regularization on the error function during training, for example, to enable regularization for individual layers of the neural network.

Efficient choice of a suitable neural network architecture is a hard and time-consuming task, so automated solutions are highly required. In recent years, a large number of related algorithms for Neural Architecture Search~(NAS) have emerged. The paper~\cite{automation} provides detailed analysis of various algorithms for automating the choice of the structure of a neural network, in particular, based on reinforcement learning and the genetic principle. A narrower case of using the choice of structure is considered in~\cite{BALDEONCALISTO202076} and~\cite{9103245} for convolutional networks. NAS in the problem of face recognition~\cite{zhu2020new} achieves~$98.77\%$ accuracy while maintaining a relatively small network size. NAS is applicable for various machine learning tasks. In~\cite{9095246}  it is used for machine translation. In~\cite{macro} a method is studied that generalizes regularized genetics and MNAS achieves~$94.46\%$ accuracy and a significant reduction in training time using the MNIST dataset as an example. The problem of computationally efficient use of NAS is studied in~\cite{CASSIMON2020100234}.
 
\section{Formulation of the problem of choosing the optimal structure}
To solve the model selection problem, we propose to construct a modified error function. We construct an error function which includes additive regularization to increase accuracy and reduce structural complexity of a deep learning model. There given a finite set of pairs
\begin{equation*}
(\mathbf{x}, y) \in \mathfrak{D}, \quad \mathbf{x} \in \mathbb{R}^{n}, \quad y \in \mathbb{R}, \tag{5}
\end{equation*}
where~$\mathbf{x}$ is a vector of independent variables,~$y$ is a dependent variable. One has to construct a structure of a model~$f$:
\begin{equation}
f(\mathbf{x})= \sigma_k\circ\underset{1\times1}{\mathbf{w}_k^\mathsf{T}\boldsymbol{\sigma}_{k-1}}\circ\mathbf{W}_{k-1}^\mathsf{T}\boldsymbol{\sigma}_{k-2}\circ\dots\circ\underset{n_2 \times 1}{\mathbf{W}_2^\mathsf{T}\boldsymbol{\sigma}_1}\circ\underset{n_1 \times n}{\mathbf{W}_1^\mathsf{T}}\underset{n \times 1}{\mathbf{x}}. \tag{6}
\end{equation}
The model structure includes a superposition of a linear model, a deep neural network, and an autoencoder.  The model parameters~$\mathbf{w}$ to optimize according to the error function is the concatenated vector
\[
\mathbf{w} = [\mathbf{w}_k, \textbf{vec}(\mathbf{W}_{k-1},\dots,\textbf{vec}(\mathbf{W}_{1})]^\mathsf{T}.
\]

The problem of choice of the optimal model structure~$\boldsymbol{\Gamma}$ is stated using the binary matrices~$\boldsymbol{\Gamma}$:
\begin{equation}\label{eq57}
f(\mathbf{x}) = \sigma_k\circ\boldsymbol{\Gamma}_k\otimes\underset{1\times1}{\mathbf{w}_k^\mathsf{T}\boldsymbol{\sigma}_{k-1}}\circ\boldsymbol{\Gamma}_{k-1}\otimes\mathbf{W}^\mathsf{T}_{k-1}\boldsymbol{\sigma}_{k-2}\circ\dots\circ\boldsymbol{\Gamma}_2\otimes\underset{n_2 \times 1}{\mathbf{W}_2^\mathsf{T}\boldsymbol{\sigma}_1}\circ\boldsymbol{\Gamma}_1\otimes\underset{n_1 \times n}{\mathbf{W}_1^\mathsf{T}}\underset{n \times 1}{\mathbf{x}}, \tag{7}
\end{equation}
where~$\boldsymbol{\Gamma}$ is the matrix defining the structure of the model;~$\otimes$ is the Hadamard product, defined as an element-by-element multiplication. If the element~$\gamma \in\{0,1\}$ of the matrix~$\boldsymbol{\Gamma}$ is equal to zero, then the corresponding element of the matrix of the parameters~$\mathbf{W}$ is zeroed and does not participate in the operation of the model. 
% The set of indices corresponding to non-zero elements of the matrix~$\Gamma$ is denoted as~$\mathcal{A}$. 
The set of non-zero elements of the matrix~$\Gamma$ forms the model structure. It is required to find model structure that provides the minimum of the function
\[
\boldsymbol{\Gamma}^*=\arg\min S\left(f_{\boldsymbol{\Gamma}} \mid \mathbf{w}^{*}, \mathfrak{D}_{\mathcal{C}}\right)
\]
on partitioning of the dataset~$\mathfrak{D}$ by the definite set of indices~$\mathcal{C}$. That is, it is required to reduce the number of features and increase the stability of the model. The optimal model parameters are
\[
\mathbf{w}^{*}=\arg\min S\left(\mathbf{w} \mid \mathfrak{D}_{\mathcal{C}}, f_{\boldsymbol{\Gamma}}\right).
\]
The algorithm for finding the optimal network structure assumes minimizing the error function.

\paragraph{Error function.} The key idea of this research paper is to construct a new error function using metaparameters of additive regularization. We  suggest a composite error function that consists of several summands. The first summand corresponds to the accuracy of approximation of the dependent variable. The second summand is the accuracy of reconstruction of the independent variable by autoencoder. The remaining~$k$ of the summands are responsible for additive regularization. A detailed discussion of the types of regularizers used is presented in Supplementary.

\begin{table}
\caption{Catalog of additive error function regularizers}\label{table:regul_table}
\centering\begin{tabular}{|l|c|}
\hline
Role in the additive regularization & Regularizer  \\
\hline\hline
Approximation error & $\|\mathbf{y}-\mathbf{f}(\mathbf{W})\|_{2}^{2}$ \\
\hline
Reconstruction error & $\|\mathbf{x}-\mathbf{r}(\mathbf{x})\|_{2}^{2}$ \\
\hline
$L_{1}, L_{2}$ regularization & $\|\mathbf{w}-\mathbf{w}_0\|_{1},\left\|\mathbf{w}-\mathbf{w}_{0}\right\|_{2}^{2}$ \\
\hline
Penalty for matrix non-orthogonality & $\left\|\mathbf{W} \mathbf{W}^{\mathsf{T}}-\mathbf{I}\right\|_{2}$ \\
\hline
Tikhonov regularization & $\|\mathbf{B W}\|_{2}$ \\
\hline
Penalty for the weights frequencies & $\|(\mathbf{I}-\mathbf{A}) \mathbf{W}\|_{2}$ \\
\hline
\end{tabular}
\end{table}

The error function includes summands~(11) and~(13) to optimize the parameters of the model~(6)
\begin{equation*}
S=\lambda_{x} E_{x}+\lambda_{y} E_{y}+\sum_{i=1}^{k} \lambda_{i} \mathcal{R}_{i}(\mathbf{W}), \tag{8}
\end{equation*}
where $\mathcal{R}_{i}=\mathcal{R}(\mathbf{W})=\left[\mathfrak{r}_{1}(\mathbf{W}), \cdots, \mathfrak{r}_{r}(\mathbf{W})\right]^{\mathsf{T}}$ is a vector consisting of the values of the regularizers of the $i$-th layer.

The metaparameters of additive regularization are collected in a matrix of $k \times r$, where $k$ is the number of layers, and $r$ is the number of regularizers in each layer:
\[
\left[\begin{array}{cccc}
\lambda_{1,1} & \lambda_{1,2} & \ldots & \lambda_{1, r}  \tag{9}\\
\ldots & \ldots & \ldots & \ldots \\
\lambda_{k, 1} & \lambda_{k, 2} & \ldots & \lambda_{k, r}
\end{array}\right].
\]
In turn, $\boldsymbol{\lambda}_{i}$ is a vector
\begin{equation*}
\boldsymbol{\lambda}_{i}=\left[\lambda_{1}, \lambda_{2}, \ldots, \lambda_{r}\right]^\mathsf{T}. \tag{10}
\end{equation*}
Each element $\lambda_{r}$ of this vector corresponds to the regularizer $\mathfrak{r}_{r}$ of the corresponding $k$-th layer. This approach allows you to vary the structure of the error function. For example, if we nullify errors $E_{y}$ and additive regularization, leaving only the error $E_{x}$ in function~(8), then the layer will behave as an autoencoder. On the contrary, with a weak regularization of the parameter $E_{x}$, an approximating layer is obtained.

When optimizing the structure $\boldsymbol{\Gamma}$ of a deep learning model, three types of quality criteria are used: accuracy, robustness, and complexity.

\paragraph{Accuracy.} For the regression problem, the error function is
\begin{equation*}
E_{y}=\sum_{(\mathbf{x}, y)\in\mathfrak{D}}\bigl(y-f(\mathbf{x})\bigr)^{2}. \tag{11}
\end{equation*}
It includes the resulting model predictions $f$ and dependent variable values $y$. Within regression, the approximation accuracy is
\begin{equation*}
\operatorname{MAE}=\frac{1}{|\mathfrak{D}|}{\sum_{(\mathbf{x}, y)\in\mathfrak{D}}|y-f(\mathbf{x})|}. \tag{12}
\end{equation*}
The principal component analysis model or autoencoder does not use dependent variables. The error function $E_{\mathbf{x}}$ penalizes the residuals of the restored input~$\mathbf{x}_{i}$ :
\begin{equation*}
E_{\mathbf{x}}=\sum_{\mathbf{x}\in\mathfrak{D}}\|\mathbf{x}-\mathbf{r}(\mathbf{x})\|_{2}^{2}, \tag{13}
\end{equation*}
where $\mathbf{r}(\mathbf{x})$ is a linear reconstruction of the object $\mathbf{x}$. The autoencoder parameters
\[
\mathbf{W}_{\mathrm{AE}}=\left\{\mathbf{W}^{\prime}, \mathbf{W}, \mathbf{b}^{\prime}, \mathbf{b}\right\}
\]
are optimized in such a way~(13) as to bring the reconstruction $\mathbf{r}(\mathbf{x})$ closer to the original vector $\mathbf{x}$.

Introduce the \emph{model robustness} as the minimum variance of the error function~(5):
\[
\operatorname{var}(S) \rightarrow \min.
\]
Introduce the \emph{structural complexity} of the model as the number of optimized parameters in~(7). It also equals to the number of non-zero values in the matrix~$\boldsymbol\Gamma$.

\paragraph{Problem statement.} The minimization problem is stated as
\begin{equation*}
\hat{\mathbf{w}}, \hat{\boldsymbol{\Gamma}}, \hat{\boldsymbol{\lambda}}=\arg \min S(\mathbf{w},  \boldsymbol{\Gamma}, \boldsymbol{\lambda}). \tag{14}
\end{equation*}
The model parameters, additive regularization metaparameters, and structure parameters are optimized to provide a minimum of the error function~(8).
\begin{figure}[!htp]
\centering\includegraphics[width=.8\textwidth]{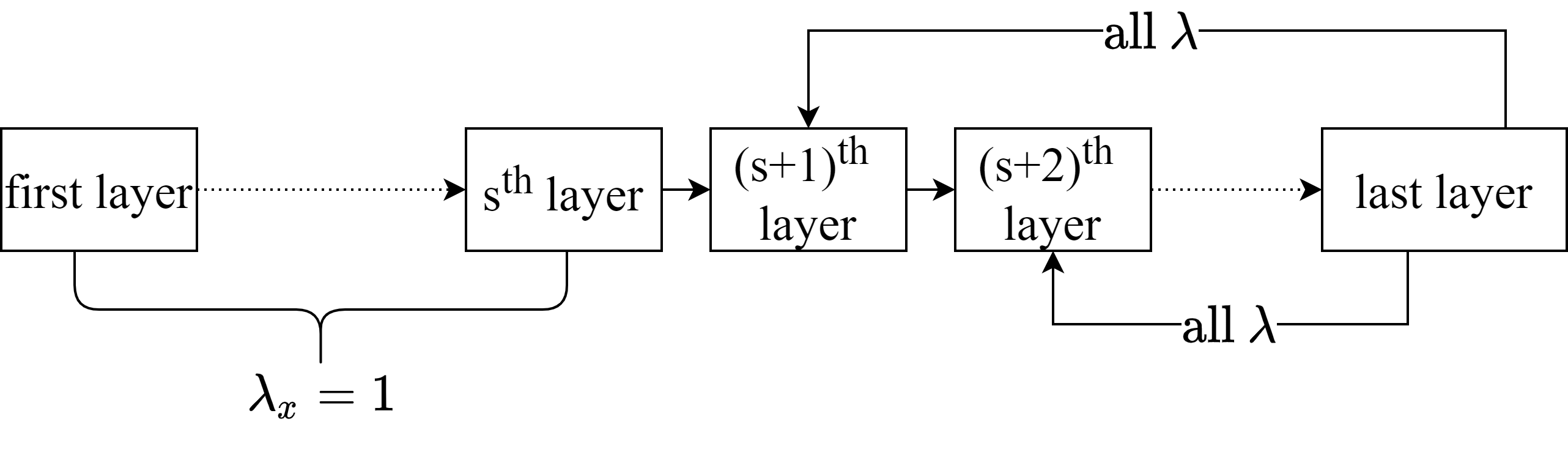}
\caption{Expert task of the optimization schedule}
\end{figure}

\section{Optimization schedule}\label{schedule} 
To solve the problem~(14), we create an optimization schedule for the metaparameters of the regularization. We have to  set the metaparameters depending on the iteration number. To set~$\boldsymbol\lambda$, we proposed two approaches: 1) Expert schedule selection, and 2) Algorithmic schedule selection.

\begin{figure}[!b]
\includegraphics[width=0.5\textwidth]{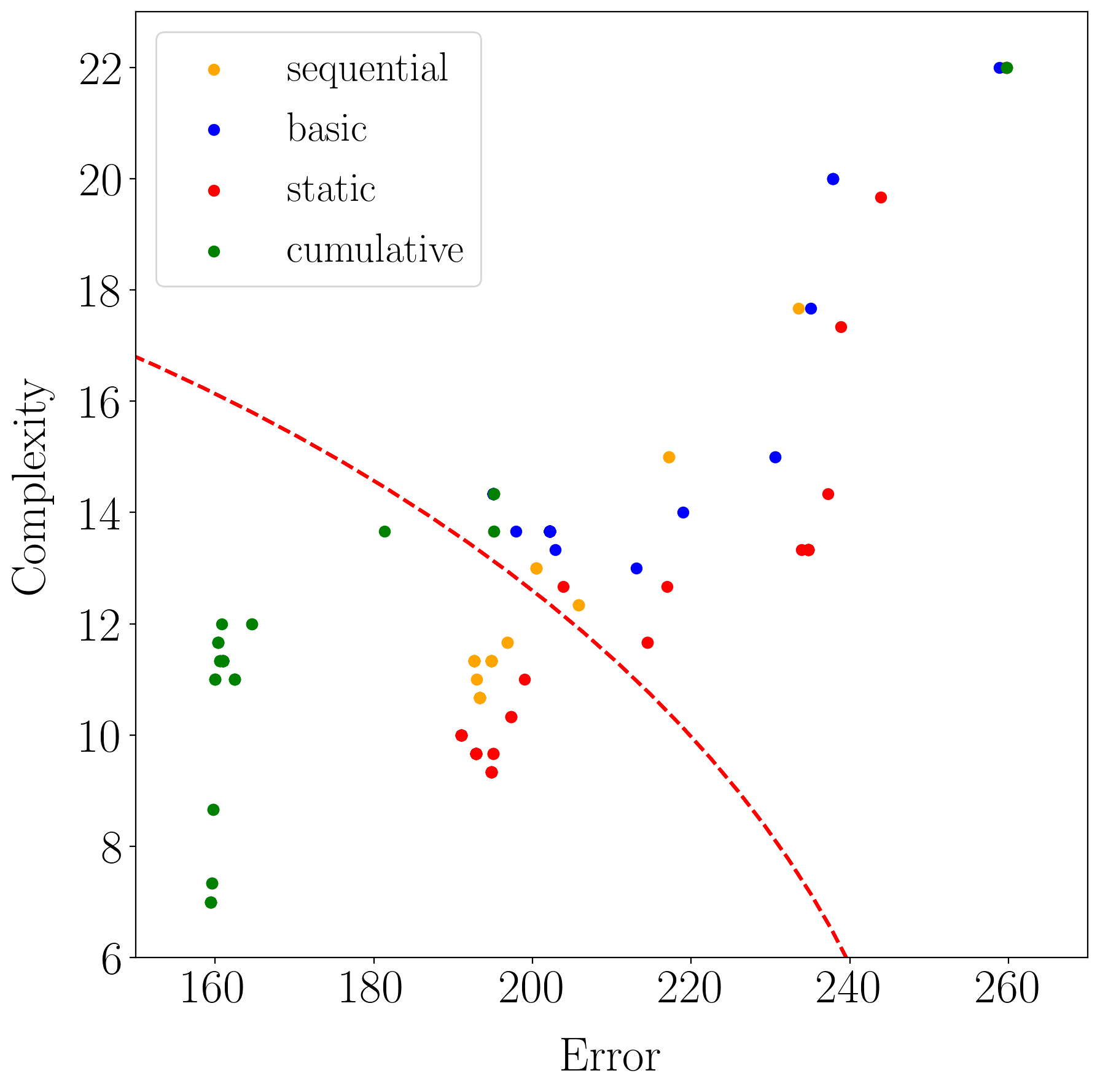}
\includegraphics[width=0.5\textwidth]{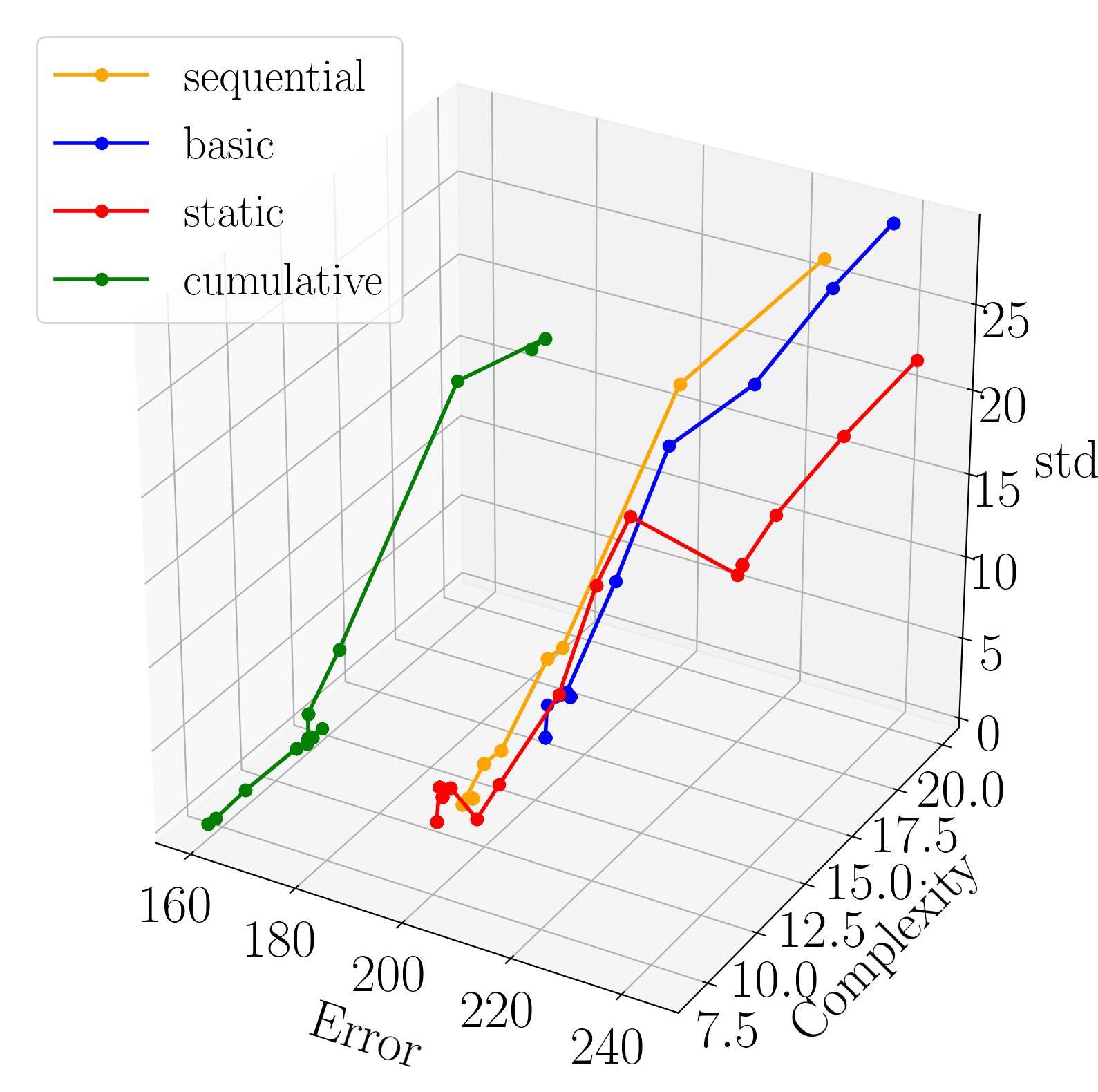}
\caption{Changes in the accuracy, complexity, and robustness of models with iterations of the GA-REG algorithm}
\label{fig:gen_3d}
\end{figure}

\paragraph{Expert schedule selection.} Let the number of layers of the neural network~(7) equals~$k$, and the number of training iterations equals~$T$.
\begin{enumerate}
\item Train $s$ first layers as autoencoders, i.e. for $s$ layers $\lambda_x = 1$ and for the last $k$  slayers $\lambda_y = 1$. The metaparameters are equal to zero, i.e. $i$ for $i = 1, \dots , s$. Optimize the autoencoder parameters at the end of the step.
\item The number of iterations $T$ is equally split between unoptimized $k-s$ layers of the neural network. At the first segment of iterations, all regularizers of $s + 1$ layer with metaparameters equals $1$ are added to~(8). On the second segment of iterations, all regularizers with parameters equal $1$ are added to the layer $s+2$, and so on.
\end{enumerate}
\paragraph{Algorithmic schedule selection.} The GA-REG is used as a method for setting up the optimization schedule. It is proposed to combine the process of optimization of neural network parameters with iterations of a genetic algorithm to optimize regularization metaparameters. The regularization metaparameters of each layer~(10) are optimized by GA-REG during $\frac{k}{T}$ iterations. The optimization procedure for each layer is:
\begin{enumerate}
\item Initiate a population of alternative metaparameters $\{\boldsymbol{\lambda}_1\dots \boldsymbol{\lambda}_p\}$.
\item For each element of this set update the neural network parameters.
\item If using one of the population vectors the error on the validation set decreases, then this vector is used as regularization metaparameters for the current layer of the neural network. If not, the current best candidate is used as the regularization weights.
\item Create a new population of regularization vectors using mutation and crossing procedures.
\item Iterate steps 2--4.
\end{enumerate}
For this optimization procedure, at every $\frac{T}{k}$ iterations, obtain the metaparameters for the current layer of the neural network. The result of the operation of the algorithm after all iterations of training the network and tuning the metaparameters is the matrix~(9).

\begin{table}[!htbp]
\caption{Description of the datasets for the experiment}\label{table1}
\centering\begin{tabular}{ | l | l | l |l | l | l | l | } \hline
Dataset $\mathfrak{D}$ & Size train  & Size val  & Size test& Objects & Features\\ \hline \hline
Protein & 27438 & 9146 & 9146 & 45730 & 9 \\ \hline
Airbnb & 6298 & 2100 & 2100 & 10498 & 16 \\ \hline
Wine quality & 2938 & 980 & 980 & 4898 & 11 \\ \hline
Synthetic & 6000 & 2000 & 2000 & 10000 & 30 \\ \hline
online news & 23786 & 7929 & 7929 & 39644 & 58 \\ \hline
conductivity & 12757 & 4253 & 4253 & 21263 & 81 \\ \hline
concrete & 618 & 206 & 206 & 1030 & 8 \\ \hline
electricity & 6000 & 2000 & 2000 & 10000 & 12 \\ \hline
NASA & 901 & 301 & 301 & 1503 & 5 \\ \hline
\end{tabular}
\end{table}

\section{Computational experiment} \label{experiment}
The procedure for optimizing the structure of a neural network and the effect of additive regularization on the quality of approximation are investigated. We have to reduce the model complexity maintaining the accuracy~(12). % the quality~\eqref{eq106} of the approximation. 
The structure of the network $\boldsymbol{\Gamma}$~\eqref{eq57} is optimized using the GA-NAS. The goal of the computational experiment is to determine the optimal values of the regularization metaparameters, as well as to study the dependence of the accuracy, complexity, and stability of the model on the regularization procedure specified by the metaparameters. The algorithm GA-REG, a procedure to construct a model is described in the previous paragraph. 

The metaparameters $\lambda$ are determined for each layer separately. The metaparameter matrix is
\[[\boldsymbol{\lambda}_1, \boldsymbol{\lambda}_2,\cdots,  \boldsymbol{\lambda}_{k}]^{\mathsf{T}},\]
the vector~$\boldsymbol{\lambda}_i$ collects metaparameters for the $i$-th layer.

\begin{figure}[!htbp]
\includegraphics[width=\textwidth]{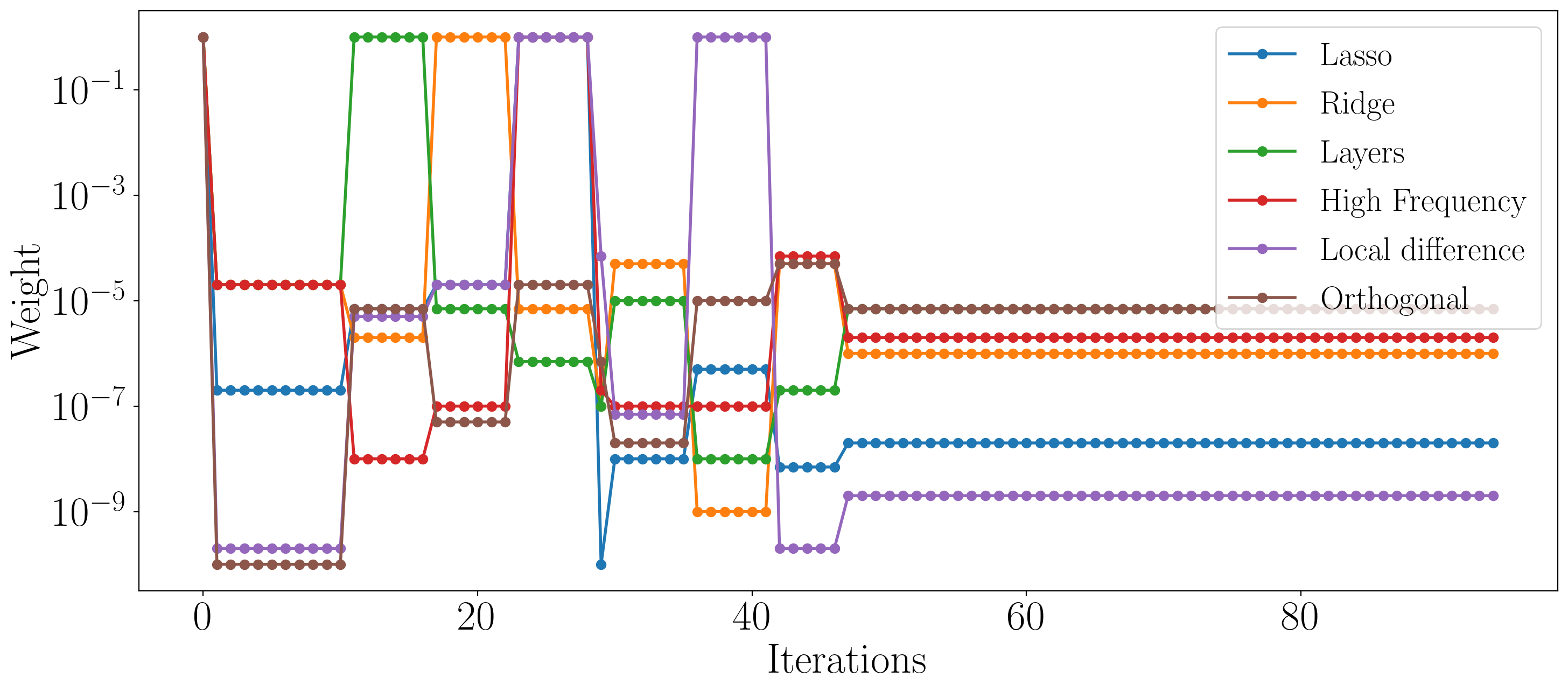}
\caption{Changing the regularization metaparameters over iterations}
\label{fig:reg_path}
\end{figure}

\paragraph{Datasets.} 
The quality of the proposed approach to constructing the error function is evaluated on several datasets from the UCI repository~\cite{Dua:2019} and one synthetic dataset. The description of all datasets is presented in Table 1. The synthetic dataset consists of features with different properties of orthogonality and correlation with each other and with the target variable. The procedure for generating synthetic data is described in~\cite {katrutsa2015stress}. Each dataset is split into three parts.
\begin{enumerate}
 \item  The training sample set includes~$60\%$ of the dataset.
 \item  The validation set includes~$20\%$ of the dataset. The vectors of regularization metaparameters are selected for this dataset. And then a genetic algorithm is applied that looks for the optimal structure. These procedures are called GA-REG and GA-NAS, respectively.
 \item The test sample set includes~$20\%$ of the original set. It does not participate in any way in optimizing the model structure and setting parameters. This set is used only for comparison of models with and without the additive regularization, for the original and optimized structures.
 \end{enumerate}

\paragraph{Algorithm implementation.} 
The main idea of the work is to develop a method for tuning regularization metaparameters, that is, the weights of each regularizer. Figure~\ref{fig:gen_3d} shows the algorithmic selection method, which is described in Section~\ref{schedule}. The legend on the charts has the following interpretation:
\begin{enumerate}
\item[1)] \textit{basic} is a model without regularization,
\item[2)] \textit{sequential} is an optimization of regularization metaparameters by GA-REG. 
The layer regularization uses the current best metaparameter vector from the genetic population at each train iteration,
\item[3)] \textit{cumulative} means that for each layer of the neural network after all iterations of the GA-REG, the best vector of regularization metaparameters is determined. This vector is used during training iterations that match the setting of the current layer. During the iterations corresponding to the setting of the first layer, the regularization metaparameters of the remaining layers are equal to zero,
\item[4)] \textit{static} means that the entire network is trained using the full regularization matrix~(9).%\eqref{matrix_lambda}.
\end{enumerate}

\begin{figure}[!htp]
\includegraphics[width=0.5\textwidth]{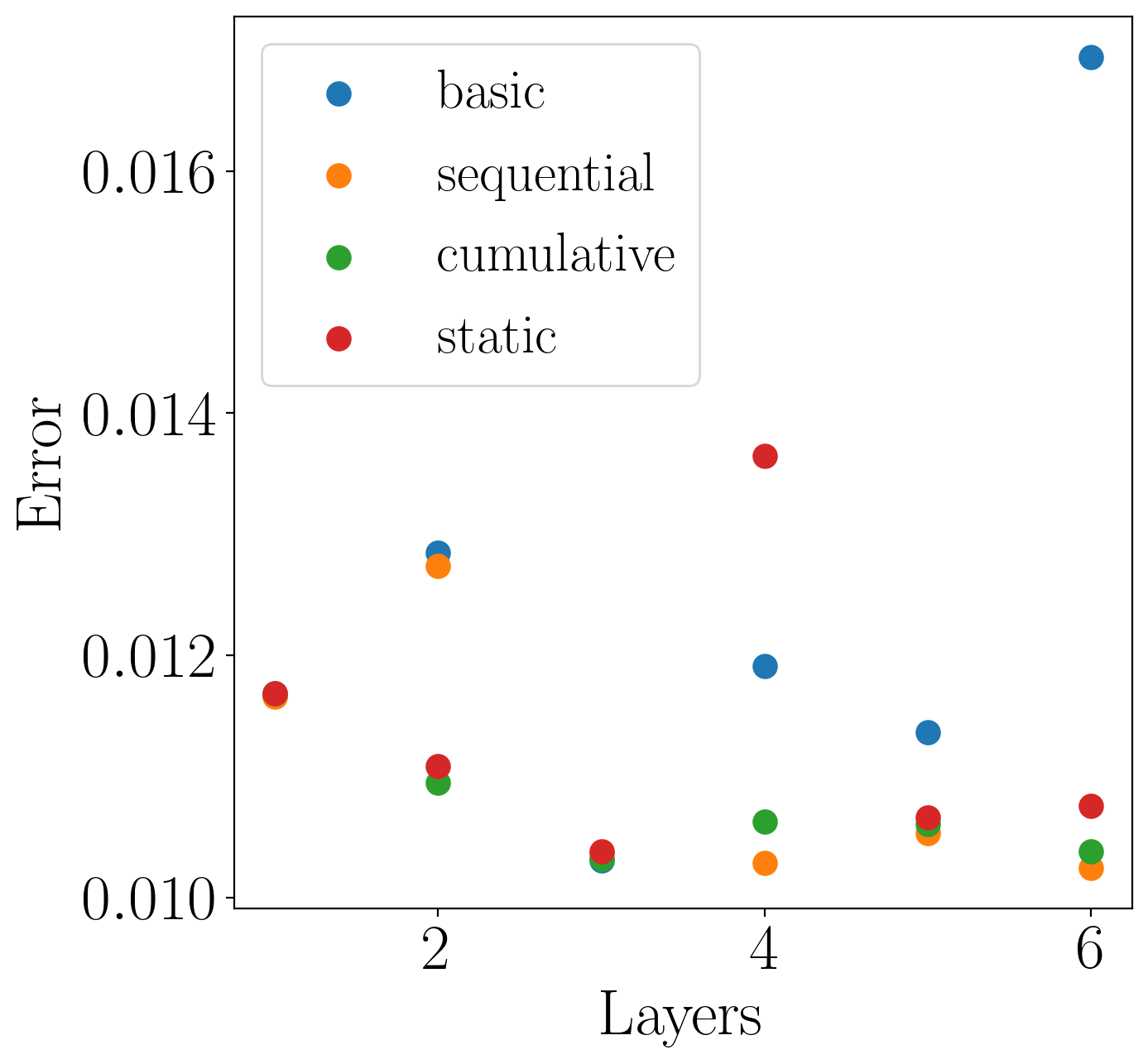}
\includegraphics[width=0.5\textwidth]{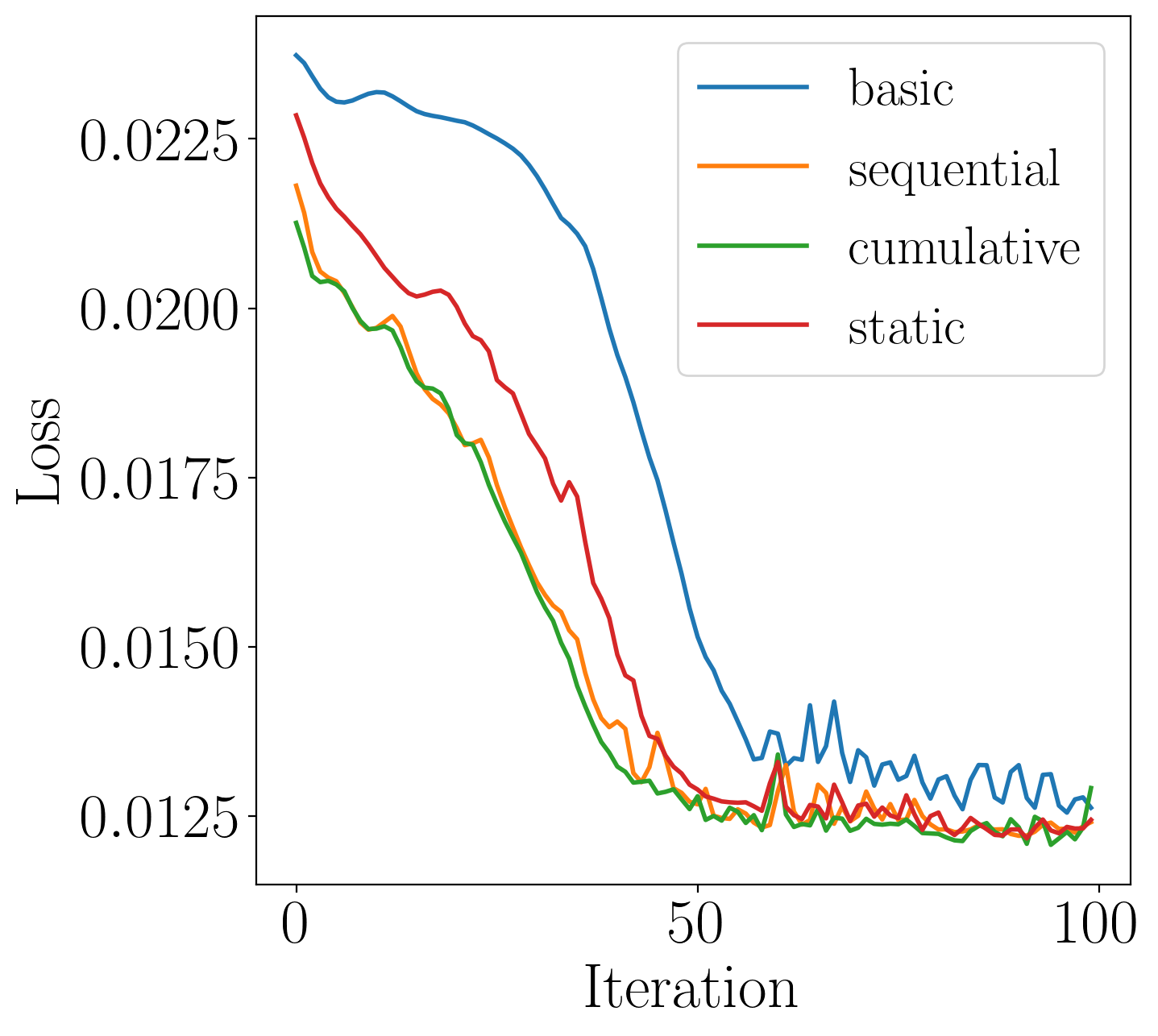}
\caption{\textit{Left:} the error~(11) %\eqref{eq:mce} 
depends on the number of layers of the neural network given the method of tuning metaparameters. \textit{Right:} the model error on the validation set  over iterations of training.}
\label{fig:qual_genetic_weights}
\end{figure}

\begin{figure}[!htbp]
\includegraphics[width=0.5\textwidth]{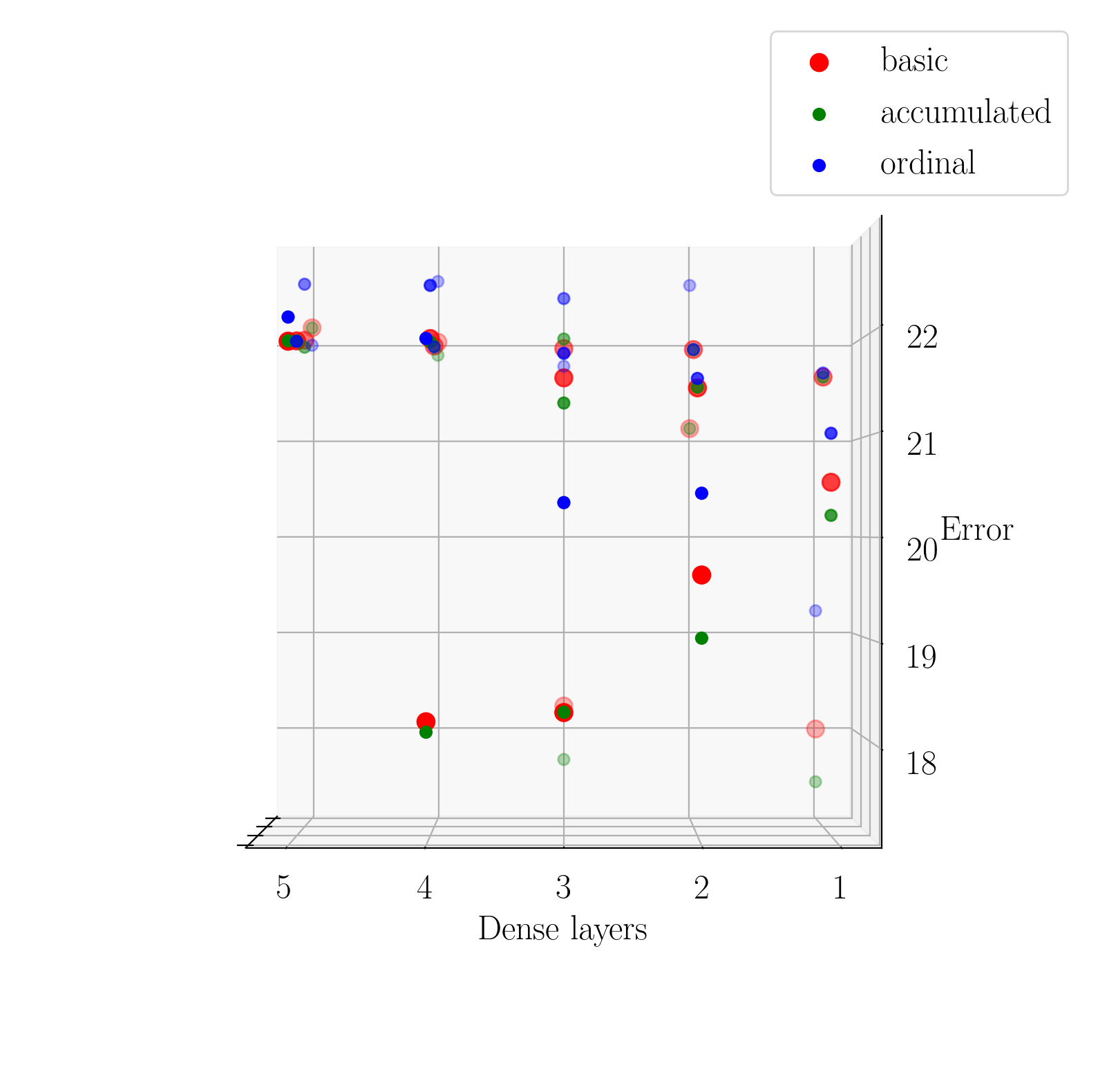}
\includegraphics[width=0.5\textwidth]{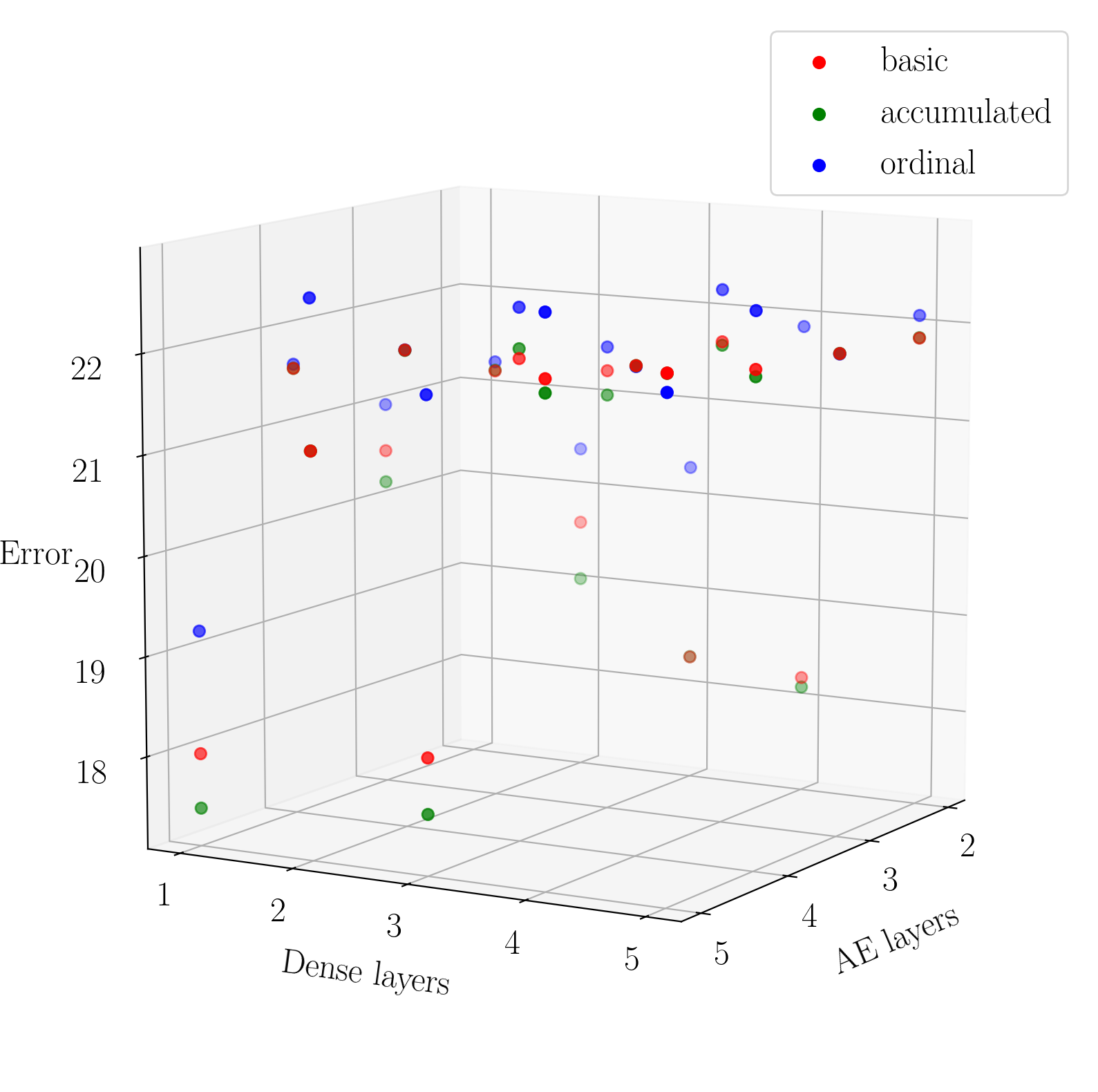}
\caption{The expert schedule; the error~(11) %\eqref{eq:mce} 
depends on the configuration of the model. \textit{Left:} a two-dimensional slice, the number of layers of a fully connected network.
\textit{Right:} Along the axis, the horizontal axes show the number of layers in the autoencoder and fully connected network, vertical axis shows the model error.}
\label{fig:qual_2_3d}
\end{figure}

To reduce the structural complexity of the neural network the GA-NAS was used. The result of the algorithm work is a binary matrix, each line of which corresponds to a binary mask. The zeros in the mask mean that the corresponding neuron in the network can be deleted and does not participate in the model optimization procedure. Thus, the GA-NAS in this case acts as a thinning procedure of a neural network, that is, a procedure for reducing the structural complexity of the model. Figure~\ref{fig:gen_3d} shows a study of the variation in accuracy, complexity, and robustness of models. The iterations of the genetic algorithm start at the top right corner of the cube of complexity-robustness-error space~(CRES) and end at the bottom left. The complexity shows the number of non-zero model neurons. It is worth noting that all regularized models are more stable than non-regularized ones, and also start with more accurate solutions. According to this figure, it can be argued that a neural network with regularized parameters converges to a less structurally complex and qualitatively more accurate solution, and also has a higher stability compared to a model that does not use additive regularization.

Table~\ref{table:table_qual} shows the model error on the test sample set, depending on the considered dataset. It is seen that in all cases the use of regularization allows one to reduce the model error. For comparison with other models that were used on the given data, the work~\cite{cortez2009modeling} was chosen. In comparison with this work, we managed to reduce the error by almost four times.
 
Figure~\ref{fig:reg_path} shows the process of adjusting the metaparameters over iterations of training by GA-REG. Each color denotes a separate regularizer used in the error function. A detailed description of regularizers is given in Table~\ref{table:regul_table} and the Supplementary section.

Figure~\ref{fig:qual_genetic_weights}\,(left) shows the dependence of the neural network error on the test sample set on the used method for setting the regularization weights and the number of neural network layers. It is seen that for any configuration of the neural network, the use of regularization allows one to obtain a smaller error. Figure~\ref{fig:qual_genetic_weights}\,(right) shows the error in the validation set during the training of the neural network. It is seen that the use of regularization leads to faster convergence and lower error.

\begin{table}[!htbp]
\caption{Model error on the test sample set depending on the considered dataset\label{table:table_qual}}
\centering\begin{tabular}{ | l | l | l |l | l | l | }
\hline
Dataset $\mathfrak{D}$ & basic  &  sequential & cumulative& static\\
\hline \hline
airbnb & 208.59\,$\pm$\,7.05& \textbf{147.22\,$\pm$\,5.22} & 168.18\,$\pm$\,4.55 & 168.18\,$\pm$\,4.55 \\
\hline
protein & 654.79\,$\pm$\,6.37& \textbf{611.14\,$\pm$\,5.9} & 643.65\,$\pm$\,6.22 & 621.69\,$\pm$\,6.23 \\
\hline
wine & 174.33\,$\pm$\,5.46& 172.35\,$\pm$\,5.43& \textbf{162.66\,$\pm$\,5.26} & 162.66\,$\pm$\,5.26	 \\
\hline
synthetic & 2.69\,$\pm$\,1.12	& 2.18\,$\pm$\,1.04& \textbf{0.39\,$\pm$\,0.26} & 0.39\,$\pm$\,0.26	 \\
\hline
conductivity & 113.62\,$\pm$\,2.42& \textbf{109.07\,$\pm$\,2.43}& 114.04\,$\pm$\,2.43 & 109.69\,$\pm$\,2.42	 \\
\hline
online news & 1.54\,$\pm$\,0.52&\textbf{1.54\,$\pm$\,0.52}& 1.54\,$\pm$\,0.52 & 1.54\,$\pm$\,0.52	 \\
\hline
concrete &244.82\,$\pm$\,18.05& 244.02\,$\pm$\,18.06&\textbf{242.12\,$\pm$\,17.6} & 243.43\,$\pm$\,17.89		 \\
\hline
electricity &	140.42\,$\pm$\,5.1	& \textbf{137.71\,$\pm$\,4.91}&140.42\,$\pm$\,5.1 & 140.42\,$\pm$\,5.1	\\
\hline
NASA &	452.75\,$\pm$\,23.8& 450.74\,$\pm$\,23.91&449.15\,$\pm$\,23.99 &\textbf{447.71\,$\pm$\,24.07}\\
\hline
\end{tabular}
\end{table}

Figure~\ref{fig:qual_2_3d}\,(left) and~\ref{fig:qual_2_3d}\,(right) show the expert method of the regularization from Section~\ref{schedule}. 
This is an easier method than algorithmic scheduling. The network configuration consists of a different number of autoencoder layers and a fully connected network; these numbers are plotted along the horizontal axes. With equal complexity, using additive regularization allows for less error than using a non-regularized network. Since the dimension of the number of layers in the autoencoder part of the network is hidden in ~\ref{fig:qual_2_3d}\,(left), the same graph is shown in 3D~\ref{fig:qual_2_3d}\,(right).

\begin{figure}[!htbp]
\includegraphics[width=\textwidth]{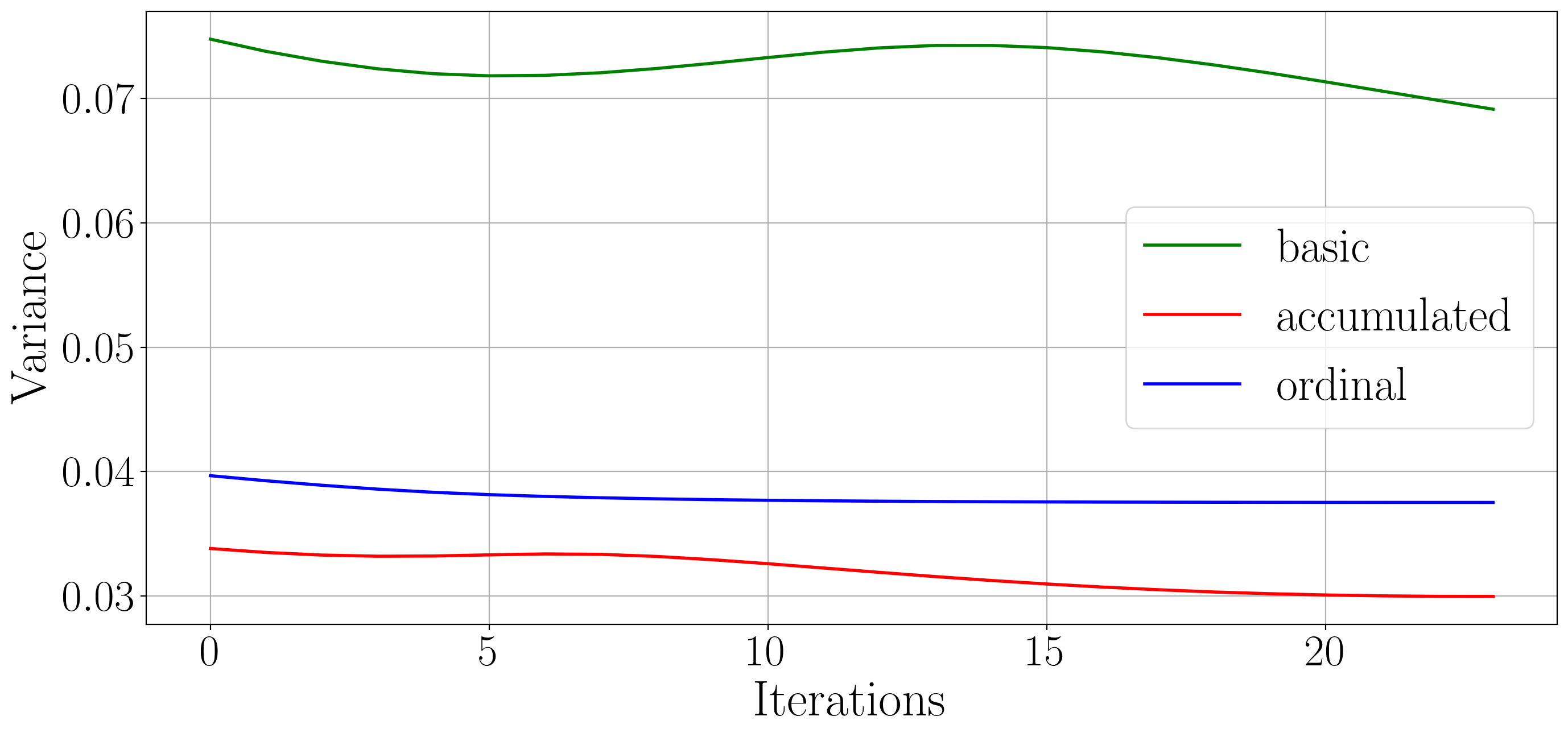}
\caption{Variation of parameter variance over iterations}
\label{fig:lines_3}
\end{figure}

Figure~\ref{fig:lines_3} shows the dependence of the variance of parameters of the model on the type of regularization used.  The \emph{basic} type in the network does not use regularization, the \emph{accumulated} type layers are sequentially included in the regularization process, the  \emph{ordinal} type as \emph{accumulated} type, only when moving to a new layer, the regularization of the previous one is turned off. It is seen that the regularized models have a lower variance of parameters.

The code of the computational experiment is available by request.% https://github.com/Intelligent-Systems-Phystech/2020-Project_Regul/

\section{Conclusion}
The research paper proposes a new method GA-REG for setting the model regularization schedule. It connects the regularization metaparameters with iterations of the training and optimizes the weight of each regularizer. It studiesthe expert and algorithmic setting of the optimization schedule. Based on the results in Table~3, the regularized models work and are more accurate than the non-regularized models. With the help of additive regularization, it was possible to achieve an increase in accuracy by~$30\%$ and a decrease in complexity by more than two times. The GA-NAS algorithm was applied to reduce the structural complexity of the model. The computational experiment shown that a neural network with correctly configured regularizers has a more preferable structure than a network without regularizers. We plan to improve our optimal structure selection method in several ways. First, we intend to generate a neural network structure by generative algorithms like autoencoder or GAN. This approach allows us to select the model structure for the specific problem. Second, sampling is a stochastic process. We plan to analyze its properties and create a more efficient generation for our tasks. Finally, we hope to develop an optimal teacher model selection method that improves the quality of knowledge distillation. 

\bibliographystyle{unsrt}
\bibliography{RegularNN2024}

\clearpage
\section*{Supplementary materials}
\subsection*{List of regularizers from the computational experiment}

\begin{enumerate}
    \item
    Lasso, $L_1$ regularization:
    \[\mathfrak{r}_1(\mathbf{w}) = \|\mathbf{w}\|_1.\]
    \item
    Layers, penalty for the number of layers:
    \[\mathfrak{r}_2(k) = k.\]
    \item
    Orthogonal, penalty for the matrix non-orthogonality:
    \[\mathfrak{r}_3(\mathbf{W}) = \|\mathbf{W}\mathbf{W}^\mathsf{T} - \mathbf{I}\|.\]
    \item 
    Variants of Tikhonov regularization:
    \begin{enumerate}
        \item
        Ridge, $L_2$ regularization:
        \[\mathfrak{r}_4(\mathbf{w}) = \|\mathbf{w}\|_2^2.\]
        \item
        High Frequency, penalty for the weight frequency:
        \[
        \mathbf{A} = \frac{1}{3} \begin{bmatrix}
            \frac{2}{3}& \frac{2}{3} & 0 & 0 &0 &0& 0\\
            1 & 1 & 1 & 1& 0 & 0 & 0 \\
            0& 0& 1 & 1& 1 & 0 & 0 \\
            0& 0& 0 & 1& 1 & 1 & 0 \\
            0& 0& 0 & 0& 1 & 1 & 1\\
            0& 0& 0 & 0 & 0& \frac{2}{3}& \frac{2}{3}
        \end{bmatrix},
        \]
        \\
        \[\mathfrak{r}_5(\mathbf{W}) =  \|(\mathbf{I} -\mathbf{A})\mathbf{W}\|.\]
        \item 
        Local difference, penalty for the local differences in weights values:
        \[
        \mathbf{B} = \begin{bmatrix}
            -2& 2 & 0 & 0 & 0 &0 &0\\
            -1 & 0 & 1 & 0& 0 & 0 & 0 \\
            0& -1& 0 & 1& 0 & 0 & 0 \\
            0& 0& -1 & 0& 1 & 0 & 0 \\
            0& 0& 0 & -1& 0 & 1 & 0\\
            0& 0& 0 & 0 & 0& -2& 2
        \end{bmatrix},
        \]
        \\
        \[\mathfrak{r}_6(\mathbf{W}) =  \|\mathbf{BW}\|.\]
    \end{enumerate}
\end{enumerate}
%GA-NAS method was described in our previous work \cite{potanin2019genetic}.

\subsection*{Theorems on the universal approximation}
Theorems 1, 2, 3 from the papers~\cite{Kolmogorov1956,Cybenko1989,math7100992} are rewritten in the notations of the present work. Below are brief version presented. 

\paragraph{Theorem 1.} Any continuous function~$f(\mathbf{x})$, defined in the $d$-dimensional unit cube could be represented as 
\[
f(\mathbf{x})=\sum_{i=1}^{2 d+1} \sigma_{i}\left(\sum_{j=1}^{d} g_{i j}\left(x_{j}\right)\right), \text { where } \mathbf{x}=\left[x_{1}, \ldots, x_{d}\right]^{\mathsf{T}}.
\]
The functions~$\sigma_{i}(\cdot), g_{i j}(\cdot)$ are continuous, and~$g_{i j}(\cdot)$ do not depend on the selection of~$f$.

\paragraph{Theorem 2.} Let~$\varphi$ be an arbitrary sigmoid function, say, 
\[\varphi(\xi) = \frac{1}{\left(1+e^{-\xi}\right)}.
\]
For any continuous real-valued function~$f$ over $[0,1]^{n}$ (or any other compact set in $\mathbb{R}^{n}$) and $\varepsilon>0$, there exist vectors~$\mathbf{w}_{1}, \mathbf{w}_{2}, \ldots, \mathbf{w}_{N}, \boldsymbol{\alpha}$ and $\boldsymbol{\theta}$, and a parametric function
\[
g(\,\cdot\,, \mathbf{W}, \boldsymbol{\alpha}, \boldsymbol{\theta}) : [0,1]^{n} \rightarrow \mathbb{R}
\] 
such that 
\[
|g(\mathbf{x}, \mathbf{W}, \alpha, \theta)-f(\mathbf{x})|<|\varepsilon|, \quad \mathbf{x} \in[0,1]^{n},
\]
where
\[
g(\mathbf{x}, \mathbf{W}, \alpha, \theta)=\sum_{i=1}^{N} \alpha_{i} \varphi\left(\mathbf{w}_{i}^{\mathsf{T}} \mathbf{x}+\theta_{i}\right)
\]
and
\[
\mathbf{w}_{i} \in \mathbb{R}^{n}, \quad \alpha_{i}, \theta_{i} \in \mathbb{R}, \quad \mathbf{W}=\left(\mathbf{w}_{1}, \mathbf{w}_{2}, \dots \mathbf{w}_{N}\right), \quad \boldsymbol{\alpha}=\left(\alpha_{1}, \alpha_{2}, \dots, \alpha_{N}\right)
\]
and
\[
\boldsymbol{\theta}=\left(\theta_{1}, \theta_{2}, \ldots, \theta_{N}\right).
\]

Before announcing Theorem 3, introduce the following notations. Use a unique activation function~$\operatorname{ReLU}$ for~$n \geqslant 1$:
\begin{equation*}
\sigma=\operatorname{ReLU}\left(x_{1}, \ldots, x_{n}\right)=\bigl(\max \left\{0, x_{1}\right\}, \ldots, \max \left\{0, x_{n}\right\}\bigr). \tag{1}
\end{equation*}
For~$n \geqslant 1$ and continuous function~$f:[0,1]^{n} \rightarrow \mathbb{R}$ introduce the norm
\[
\|f\|_{C_{0}}:=\sup _{\mathbf{x} \in[0,1]^{n}}|f(\mathbf{x})|.
\]
Use the modulus of continuity 
\[
\omega_{f}(\varepsilon):=\sup \{|f(\mathbf{x})-f(\mathbf{y})|:|\mathbf{x}-\mathbf{y}| \leqslant \varepsilon\}.
\]
Denote by~$D_{1}$ a set of any admissible neural networks with input variable in $\mathbb{R}^n$ and hidden layers in $\mathbb{R}^s$, that approximate any continuous function~$f$ that is positive-valued and defined on the unit cube~$[0,1]^{n}$. Define the minimal dimensionality of hidden layers of the neural network with the input size~$n$ from~ $D_{1}$ as
\[
\omega_{\min }(n):=\min _{s \in D_{1}}\{s\}.
\]

In a similar war, denote by~$D_{2}$ a set of any admissible neural networks with input variable in $\mathbb{R}^n$ and hidden layers in $\mathbb{R}^s$, that approximate any continuous function~$f$ that is positive-valued and defined on the unit cube~$[0,1]^{n}$. Define the minimal dimensionality of hidden layers of the neural network with the input size~$n$ from~ $D_{2}$ as
\[
\omega_{\min }^{\text {conv }}(n):=\min _{s \in D_{2}}\{s\}.
\]

\paragraph{Theorem 3.} Let $n \geqslant 1$ and $f:[0,1]^{n} \rightarrow \mathbb{R}_{+}$ is positive-valued function with norm~$\|f\|_{C_{0}}=1$. Then

\emph{Part 1.} 
If $f$ is continuous, there exists a sequence of forward-propagation neural networks~$\mathcal{N}_{k}$ with activation functions~(1),  with input size~$n$, hidden size~$n+2$, output size 1 (scalar) such that 
\begin{equation*}
\lim _{k \rightarrow \infty}\left\|f-f_{\mathcal{N}_{k}}\right\|_{C^{0}}=0. \tag{2}
\end{equation*}

In particular, $\omega_{\min }(n) \leqslant n+2$.  if one fixes $\varepsilon>0$ and compute the modulus of continuality of the function~$f \omega_{f}(\varepsilon)$, the there exists a feed-forward neural network~$\mathcal{N}_{\varepsilon}$ with activation function ReLu (1), input size $n$, hidden size $-n+3$, and output size $1$ and
\[
\operatorname{depth}\left(\mathcal{N}_{\varepsilon}\right)=\frac{2 \cdot n!}{\omega_{f}(\varepsilon)^{n}}
\]
such that
\[
\left\|f-f_{\mathcal{N}_{\varepsilon}}\right\|_{C_{0}} \leqslant \varepsilon.
\]

\emph{Part 2.} 
If $f$ is convex then there exists a sequence of feed-forward neural networks $\mathcal{N}_{m}$ wit activation functions $\operatorname{Re} L U$ (1), input size $n$, hidden size $-n+1$, and output size $1$ such that
\begin{equation*}
\lim _{m \rightarrow \infty}\left\|f-f_{\mathcal{N}_{m}}\right\|_{C_{0}}=0. \tag{3}
\end{equation*}
In particular, it holds $\omega_{\min }^{\mathrm{conv}}(n) \leqslant n+1$. Also, there exists $C>0$ such that is $f$ convex and continuous with the Lipshitz constant $L$ then neural networks $\mathcal{N}_{m}$ from (3) could be chosen so that it holds 
\[
\operatorname{depth}\left(\mathcal{N}_{m}\right)=m,\left\|f-f_{\mathcal{N}_{m}}\right\|_{C_{0}} \leqslant C L n^{\frac{3}{2}} m^{-\frac{2}{n}}.
\]

\emph{Part 3.} 
If  $f$ is continous then there exists a constant $K$ that depend on $n$ only and a constant $B$ that depends on  maximum $K$ derivative of  $f$ such that for any $m \geqslant 3$ the neural networks  $\mathcal{N}_{m}$ with size $n+2$ from (2) could be chosen such that
\[
\operatorname{depth}\left(\mathcal{N}_{m}\right)=m,\left\|f-f_{\mathcal{N}_{m}}\right\|_{C_{0}} \leqslant B(m-2)^{-\frac{1}{n}}.
\]
\end{document}